\pdfoutput=1

\documentclass[11pt]{article}

\usepackage[preprint]{acl}

\usepackage{times}
\usepackage{latexsym}
\usepackage{enumitem}
\usepackage{amsmath}
\usepackage{amsfonts}
\usepackage{adjustbox}
\usepackage{multirow}
\usepackage{listings}
\renewcommand{\lstlistingname}{Algorithm}

\usepackage{algorithm}
\usepackage{algorithmic}

\usepackage[T1]{fontenc}

\usepackage[utf8]{inputenc}

\usepackage{microtype}

\usepackage{inconsolata}

\usepackage{graphicx}
\usepackage{etoc}
\usepackage{tcolorbox}
\tcbuselibrary{breakable}
\usepackage{booktabs}
\usepackage{todonotes}
\tcbuselibrary{listingsutf8}
\usepackage{listings}
\usepackage{longtable}
\usepackage{booktabs}
\usepackage{array}
\usepackage{xcolor}
\usepackage{subcaption}
\usepackage{float}

\definecolor{ocr}{HTML}{00C8FF}
\definecolor{ocr}{HTML}{009900}
\definecolor{joeColor}{rgb}{0.6, 0.2, 1.0}

\newcommand{\joe}[1]{\todo[color=joeColor!20,size=scriptsize,fancyline]{\textbf{Joe: } #1}}

\title{Text or Pixels? It Takes Half: \\ On the Token Efficiency of Visual Text Inputs in Multimodal LLMs}

\author{Yanhong Li\thanks{These authors contributed equally to this work.} \\
  Allen Institute for AI \\
  \texttt{yanhongl@allenai.org} \\\And
  \hspace{-1.0cm}Zixuan Lan$^{*}$ \\
  \hspace{-1.0cm}University of Chicago \\
  \hspace{-1.0cm}\texttt{zixuanlan@uchicago.edu} \\\And
  \hspace{-0.5cm}Jiawei Zhou \\
  \hspace{-0.5cm}Stony Brook University \\
\hspace{-0.5cm}\texttt{jiawei.zhou.1@stonybrook.edu} }

\begin{document}
\maketitle
\pagestyle{empty}

\begin{abstract}

Large language models (LLMs) and their multimodal variants can now process visual inputs, including images of text. This raises an intriguing question: \emph{can we compress textual inputs by feeding them as images to reduce token usage while preserving performance}?
In this paper, we show that visual text representations are a practical and surprisingly effective form of input compression for decoder LLMs.
We exploit the idea of rendering long text inputs as a single image and provide it directly to the model. This leads to dramatically reduced number of decoder tokens required, offering a new form of input compression. Through experiments on two distinct benchmarks—\textsc{RULER} (long-context retrieval) and CNN/DailyMail (document summarization)—we demonstrate that this \emph{text-as-image} method yields substantial token savings (often nearly half) without degrading task performance.\footnote{Code is available at \url{https://github.com/yanhong-lbh/text_or_pixels}.}

\end{abstract}

\section{Introduction}

Running large language model (LLM) inference on long text inputs is computationally expensive due to the underlying architecture of the Transformer model, where the self-attention mechanism's complexity scales quadratically with the input length~\citep{vaswani2017attention}. This can be prohibitive when processing long documents \citep{liu2023lost}, interactive dialogues \citep{zhou2022online}, or complex multi-step reasoning \citep{feng2025unraveling}. Even with recent increases in context length, deploying LLMs at scale (e.g., in chat assistants or document analysis) is constrained by throughput and cost per token~\citep{palm540b2022,duoattention2024,refreshkv2025, li2025chunk}. Reducing the token length of inputs without losing information is therefore highly desirable for improving LLM efficiency~\citep{recycledattention2024,tokencarve2025,vist2025, li-etal-2025-context}.

\begin{figure}[t]
    \centering
    \includegraphics[width=1\linewidth]{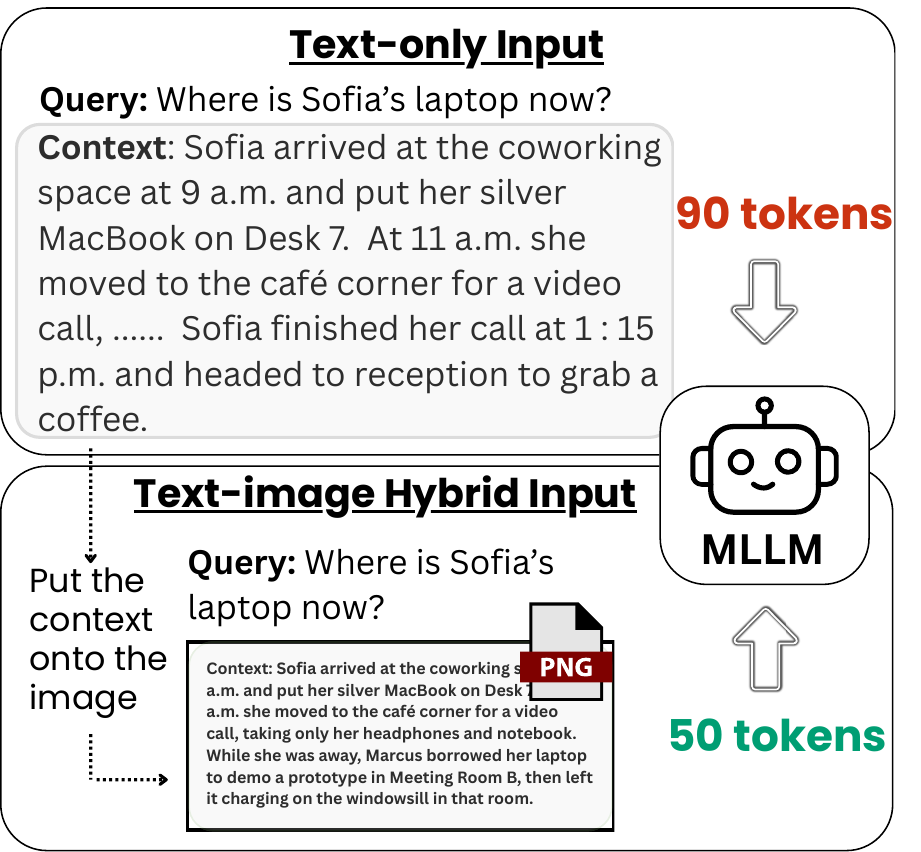}
    \caption{Illustration of our \emph{text-as-image} compression pipeline. Instead of feeding the entire 90-token context to the model (top), we convert the context into a single image and supply only the textual query alongside the image (bottom). The multimodal LLM (MLLM) reads the context from the image, so it processes just 50 visual tokens as input to the LLM decoder—cutting token usage by nearly half while still providing all the information needed to answer the question.}
    \label{fig:img_pipeline}
\end{figure}

One novel avenue for input compression is to leverage the ability of multimodal LLMs (MLLMs) \citep{liu2023visual, fang2024uncertainty} to read text from images. The adage ``a picture is worth a thousand words'' hints that visual representations might convey the equivalent of many text tokens in a compact form. Modern multimodal models like GPT-4 Vision \citep{openai2023gpt4} and Google Gemini 2.5 \citep{comanici2025gemini25pushingfrontier} can accept images as part of their input and reason over them. This raises the question: \textit{Can we feed an LLM an image of text in lieu of the text itself, to save input tokens and still get the correct output?} Early explorations (discussed in Section~\ref{sec:related}) suggest that multimodal LLMs can interpret text-based images, but the impact on efficiency and downstream performance remains under-examined.

In this paper, we present an empirical study of multimodal LLMs that explores using visual text inputs as a form of input compression. By rendering long passages as a single image, the vision encoder produces a compact set of visual tokens for the decoder, directly reducing sequence length without fine-tuning or supervision. On the \textsc{RULER} needle-in-a-haystack task, GPT-4.1-mini and Qwen2.5-VL-72B sustain $97$–$99\%$ accuracy with up to 58\% fewer decoder tokens, and on CNN/DailyMail summarization, this approach \emph{outperforms} two specialized pruning baselines at matched or higher compression rates. Although vision encoding adds some overhead on smaller models, the shorter decoder sequence yields up to 45\% end-to-end speedup on larger ones, demonstrating that off-the-shelf multimodal LLMs can treat images as an implicit \emph{compression layer}, preserving performance at nearly \textbf{half} the original text-token cost.

\section{Related Work}\label{sec:related}

\paragraph{Multimodal Variations of LLMs}

Recent advances in LLMs have extended their capabilities beyond text to handle images \citep{liu2023visual}, speech \citep{wang2023slm}, and video \citep{tang2025video}.
In the visual domain, numerous vision–language models (VLMs) have been developed, 
including BLIP(-2) \citep{li2022blip, li2023blip}, Flamingo \citep{alayrac2022flamingo}, LLaVA \citep{liu2023visual}, InternVL \citep{chen2024internvl}, Qwen-VL \citep{wang2024qwen2}, PaLI-Gemma \citep{beyer2024paligemma}, and proprietary systems such as GPT-4V \citep{achiam2023gpt} and Gemini \citep{team2023gemini}. A widely adopted architecture comprises a \textit{vision encoder} that extracts image features, a \textit{projection layer} that maps these features into the LLM’s token space, and a \textit{text decoder} that jointly processes visual and textual tokens \citep{liu2023visual, beyer2024paligemma, wang2024qwen2}. Importantly, the number of visual tokens produced by this pipeline is usually small—constrained by image size and model design—and can be far fewer than the text tokens needed to represent the same information. This suggests that multimodal LLMs already embody a form of token compression, motivating our exploration of representing rich texts directly as images to reduce decoder token usage.

\paragraph{Text as Image}
Several works have explored the idea of providing text to LLMs via images. \citet{lyu2025pixelworldperceivingpixels} proposed a pixel-input benchmark, finding that some advanced VLMs can interpret and reason over text in images, though performance can drop without specialized training. Others, such as \citet{lu2024textpixeladvancinglongcontext}, have investigated representing large documents as visual patches for handling extended context windows. While these efforts highlight the feasibility of \textit{text-as-image} inputs, they have not systematically evaluated the trade-off between token usage and reasoning performance on multi-step tasks.

Our work also relates to the concept of \emph{hybrid text-vision} prompting, where instructions are given partially in text and partially as an image \citep{aggarwal2025programmingpixelscomputerusemeets}. However, prior studies focus more on the novelty of multimodal usage or coding from screenshots, rather than on compression. We instead emphasize the efficiency gains and cost savings arising from replacing a sizeable chunk of text with an image for advanced reasoning tasks.

\paragraph{Information Compression} There are many works focusing on context compression \citep{pradeep2023rankvicuna, xu2024recomp, jiang-etal-2024-longllmlingua, li-etal-2025-context}. Our approach is complementary to \emph{soft prompt} line of work. Extreme Retrieval-Augmented Generation (xRAG) replaces full documents with one dense embedding token, achieving a $50\times$ compression ratio without fine-tuning the LM~\citep{cheng2024xrag}.
Instruction-Aware Contextual Compression (IACC) filters noisy RAG passages based on the user query, halving context length while retaining QA accuracy~\citep{hou2024iacc}.
Prompt-centric surveys catalogue a spectrum of hard pruning, abstractive summarization, and learned soft tokens that collectively reach $5$–$10\times$ compression on reasoning benches~\citep{li2024promptcompress,jha2024characterizing}.
Broader reviews on extending LLM context windows emphasize that modality fusion and token dropping are orthogonal, and can be stacked for additive gains~\citep{wang2024limitssurveytechniquesextend,li2024retrievalaugmentedgenerationlongcontext}. Distinct from these token-level approaches, we compress entire text spans by offloading them to the vision encoder of an off-the-shelf multimodal LLM—treating thousands of tokens as a single image, usually represented with fixed amounts of visual tokens or proportional to the image resolutions—and thus reduce context length to the the LLM decoder without any model fine-tuning.

\begin{figure*}[h!]
    \centering
    \begin{subfigure}[b]{0.49\textwidth}
        \centering
        \includegraphics[width=\textwidth]{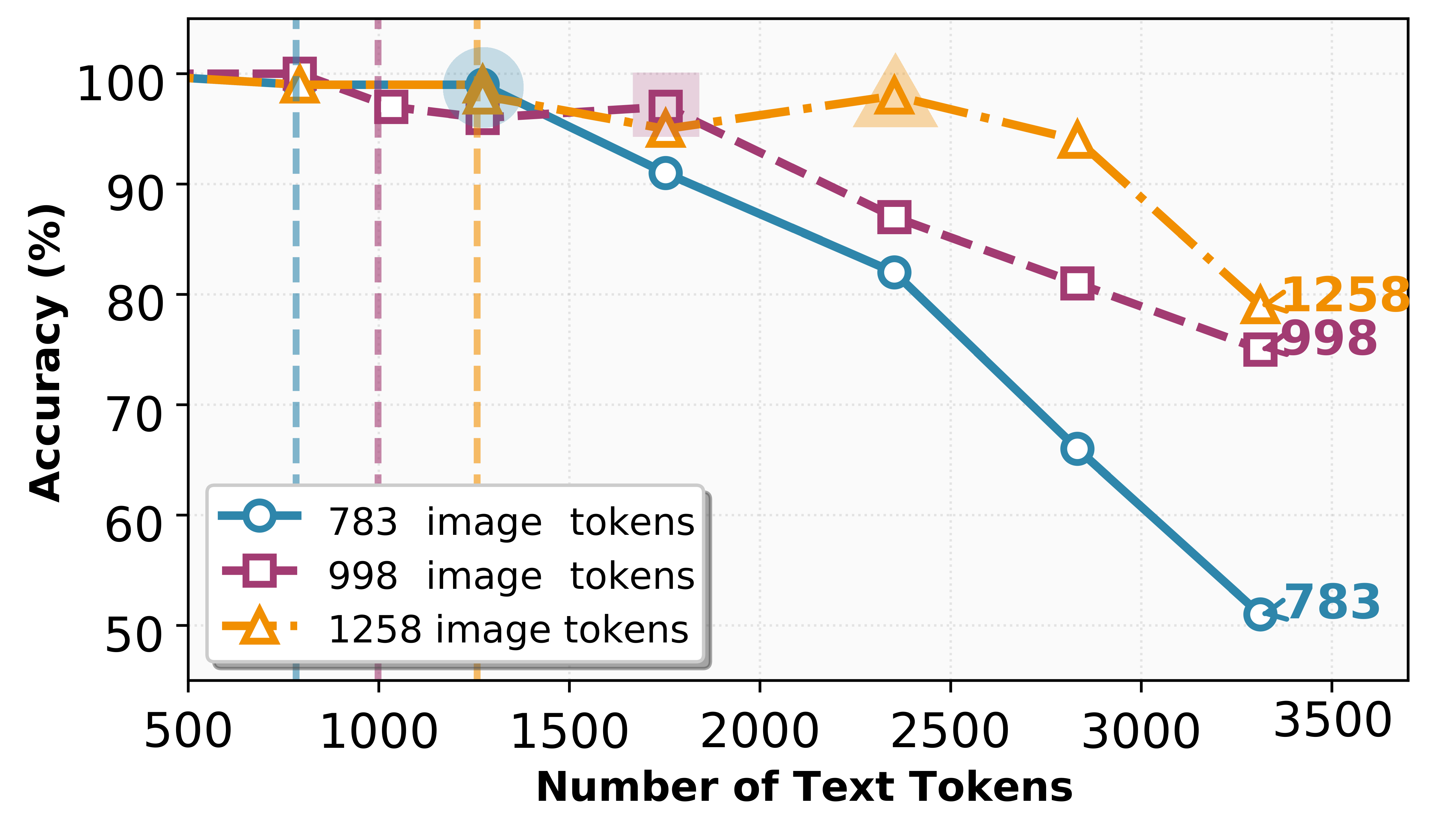}
        \caption{GPT-4.1-mini}
        \label{fig:gpt41_curve_sub}
    \end{subfigure}
    \begin{subfigure}[b]{0.49\textwidth}
        \centering
        \includegraphics[width=\textwidth]{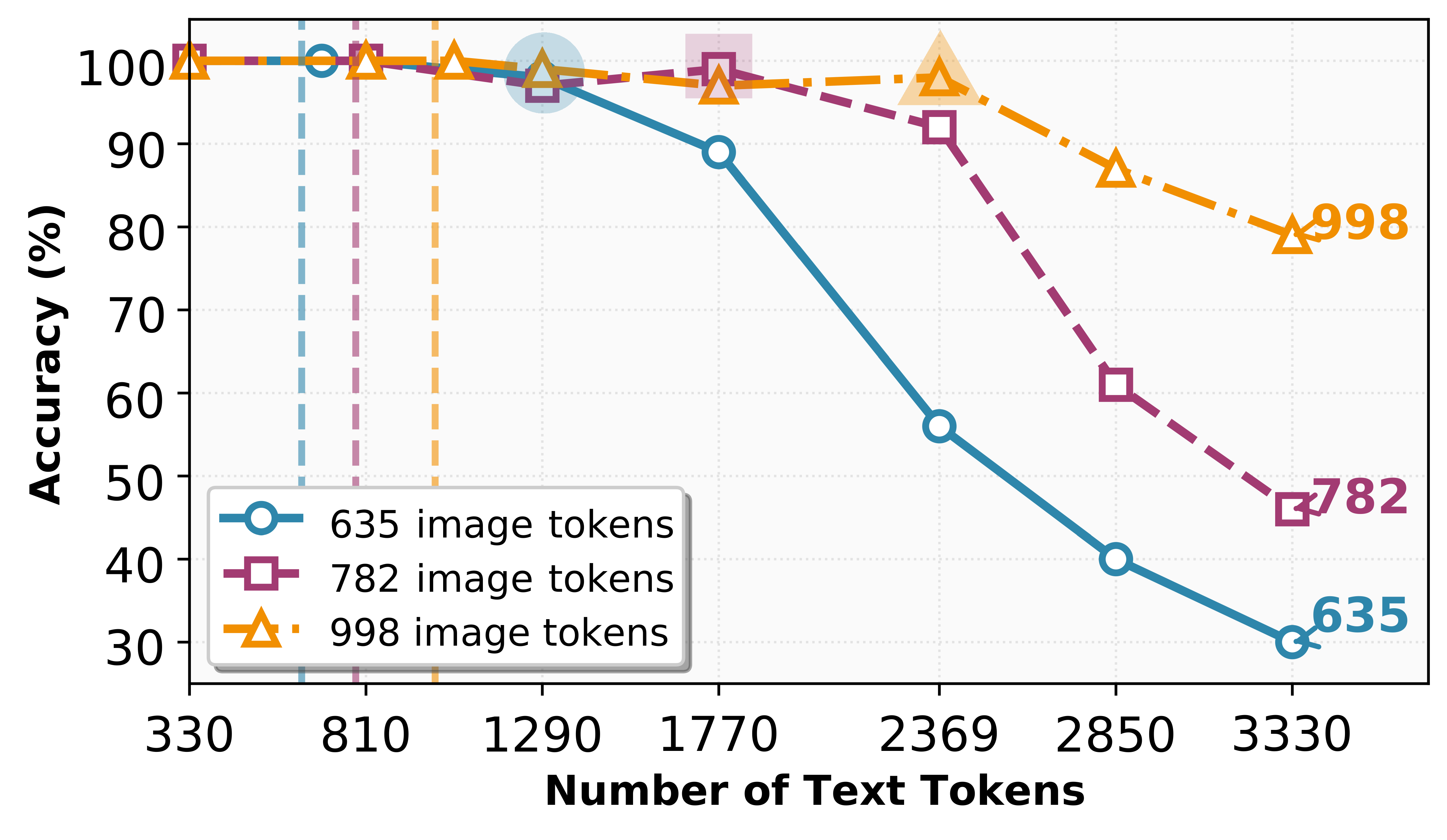}
        \caption{Qwen2.5-VL-72B-instruct}
        \label{fig:qwen_curve_sub}
    \end{subfigure}
    \caption{\textbf{Accuracy vs. text-token size ($m$) with fixed visual tokens ($k$)} for GPT-4.1-mini (left) and Qwen2.5-VL-72B-Instruct (right). Each curve varies the rendered text length at a fixed $k$; vertical dashed lines mark $m=k$, where text and visual tokens are equal. Both models degrade as text density increases, though Qwen’s larger decoder sustains higher ratios before a sharper drop. The \emph{text-token tolerance} (largest $m$ within 3 points of the text-only baseline) is shaded in the plots and reported in Table~\ref{tab:nihs_accuracy}. Beyond these limits, accuracy falls rapidly, revealing the maximum achievable compression without loss.}
    \label{fig:combined_curves}

\end{figure*}

\section{Methodology}
\label{sec:method}

\subsection{Problem Formulation}

Let the \emph{context} (e.g., a document or multi-turn dialogue) be a sequence of $m$ text tokens,
$
\mathbf{c} = (t_1, t_2, \ldots, t_m)
$,
and let the accompanying question be a (usually short) sequence
$
\mathbf{q} = (q_1, q_2, \ldots, q_{|\mathbf{q}|})\,.
$

\paragraph{Text-only baseline.}
In the standard setting we feed the concatenated token sequence
$
\mathbf{s}_{\text{text}} = (\mathbf{c},\,\mathbf{q})
$
to a \emph{text-only} LLM.  
The input‐length budget is therefore
$
T_{\text{text}} = m + |\mathbf{q}|.
$

\paragraph{Text–image (hybrid) input.}
To compress long context, we first render it into an image via
$
I = \mathcal{R}\bigl(\mathbf{c}\bigr),
$
using a LaTeX‐based typesetting pipeline that preserves the layout and line‐breaks of the original text (see Figure~\ref{fig:img_pipeline}).  
A frozen vision encoder\footnote{For all experiments we use the native vision module shipped with each multimodal LLM. No fine-tuning or additional supervision is applied.}  
\[
\Phi : I \longmapsto (v_1, v_2, \ldots, v_k), \qquad v_j \in \mathbb{R}^d,
\]
transforms the image into a \emph{fixed-length} sequence of $k$ \textit{visual tokens}.  
These embeddings are passed through a projection layer $\psi$ (e.g., a linear map) and become part of the language decoder’s input:\footnote{The number of visual tokens $k$ passed to the language decoder depends on the VLM architecture: some models (e.g., LLaVA \citep{liu2023visual}) output a fixed number, while others (e.g., Qwen-VL \citep{wang2024qwen2}) vary with image sizes. In our experiments, for Qwen models, we define $k$ as the number of visual embeddings passed to the text decoder. For GPT-4.1-mini, we use the input token count returned by the API, which accounts for the visual input.}
\[
\mathbf{s}_{\text{img}} = \bigl(\psi(v_1),\ldots,\psi(v_k),\,\mathbf{q}\bigr).
\]
The corresponding token budget is
\[
T_{\text{img}} = k + |\mathbf{q}|,\quad\text{with}\;k\!\ll\! m\;\text{possible in practice}.
\]

\noindent\textbf{Compression ratio.}
We define the \emph{compression ratio}
\[
\rho = \frac{T_{\text{text}}}{T_{\text{img}}}
      = \frac{m + |\mathbf{q}|}{k + |\mathbf{q}|}
      \approx \frac{m}{k}\quad\text{when }m\,\&\,k\gg|\mathbf{q}|.
\]
A higher $\rho$ indicates greater token savings.

\subsection{Evaluation Protocol}

For each example $(\mathbf{c},\mathbf{q})$ we run both modes: (1) \textbf{Text-only:} Evaluate the LLM on $\mathbf{s}_{\text{text}}$ to obtain answer $a_{\text{text}}$; (2) \textbf{Hybrid:} Evaluate the multimodal LLM on $\mathbf{s}_{\text{img}}$ to obtain answer $a_{\text{img}}$.
We measure \textbf{accuracy} on task-specific metrics; \textbf{token usage} $(T_{\text{text}},T_{\text{img}})$ and thus $\rho$; and \textbf{throughput and latency} (wall-clock time per example). Unless stated otherwise, $k$ is fixed by the vision encoder, so any reduction in $m$ directly translates to lower LLM decoder token cost.

\section{Experiments and Results}
\label{sec:results}

We test to what extent visual inputs of texts can reduce discrete token consumption in LLM decoders without performance degradation.
Long context tasks are especially targeted, including information retrieval and summarization, with configurable context lengths.
We test two prominent multimodal LLMs, the proprietary GPT-4.1-mini \citep{openai_gpt41mini} and the open-weight Qwen2.5-VL-72B-Instruct \citep{bai2025qwen2}.\footnote{Experiments on a smaller model Qwen2.5-VL-7B are also presented in Figure~\ref{fig:token_tolerance_scaling} and Appendix~\ref{appendix:qwen-7b}.} 

All text-as-image rendering is performed with \texttt{pdflatex} followed by rasterization at 300~dpi. Algorithmic details can be found in Appendix~\ref{sec:ConTexImage}.
During inference, we use the \texttt{temperature=0} setting for deterministic outputs and truncate responses to the first newline to obtain concise answers.

\subsection{Long-Context Retrieval}
\label{subsec:long-context-retrieval}

\paragraph{Setup.}

We evaluate our text-as-image compression strategy on the \textsc{RULER} S-NIAH (single needle-in-a-haystack) task \citep{hsieh2024rulerwhatsrealcontext}, where a single target number (\emph{needle}) is hidden in a long distractor passage (\emph{haystack}). The model must return the exact number, testing long-context retention. For each model, we generate 100 random passages and report \emph{accuracy} (percentage of correct extractions). Since the query $\mathbf{q}$ is short, the effective token budgets simplify to $T_{\text{text}}=m$ and $T_{\text{img}}=k$, giving compression ratio $\rho=m/k$.

\paragraph{How much can we compress?}

Figures~\ref{fig:combined_curves} sweep $m$ while holding $k$ fixed. Accuracy remains stable until a critical point $m^\star(k)$, after which it drops sharply. We define $m^\star$ as the largest $m$ within three percentage points of the text-only baseline, referring to this threshold as the \textit{text-token tolerance}. These values are highlighted in the plots (shaded) and reported in Table~\ref{tab:nihs_accuracy}. For example, at $k=783$, GPT-4.1-mini tolerates $m^\star\!\approx\!1{,}300$ tokens—equivalent to $\rho\!\approx\!1.9$ compression—\emph{without measurable degradation}. Larger visual budgets ($k=998,1{,}258$) increase tolerance to over 2{,}300 tokens while still saving 42–58\% of the decoder context.\footnote{Some visual examples and details in Appendix~\ref{sec:Images}.} Qwen2.5-VL follows the same trend but exhibits a steeper accuracy drop once $m^\star$ is exceeded, underscoring that text-token tolerance is both model- and $k$-dependent.

\begin{figure}[t]
\centering
\includegraphics[width=\linewidth]{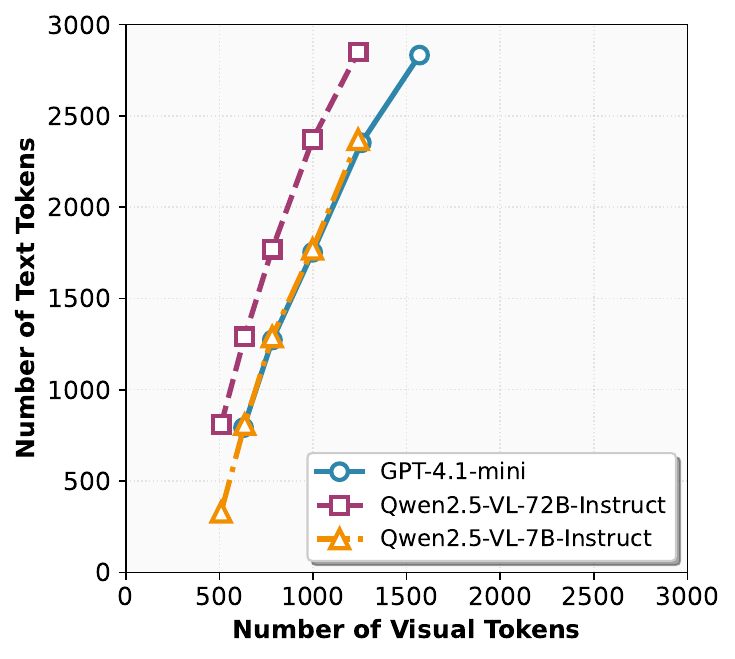}
\caption{\textbf{Text-token tolerance vs. visual token count.} The maximum text tokens $m^\star$ that can be preserved without accuracy loss, plotted against the visual tokens $k$ generated from the image. Results show a consistent reduction of roughly $1/2$ in decoder tokens.}
\label{fig:token_tolerance_scaling}
\end{figure}

\paragraph{Token savings and Latency analysis.}
Table~\ref{tab:nihs_accuracy} collates $(k,m^\star)$ and confirms that
hybrid prompts cut the decoder context \textbf{nearly in half} while matching
text-only accuracy across both models.\footnote{Additional results on a different long-context reasoning task with Gemini (Appendix~\ref{app:babilong}) also confirm this observation.}
To better understand this relationship, Figure~\ref{fig:token_tolerance_scaling} plots the text-token tolerance $m^\star$ as a function of the visual token budget $k$. For all models tested, there is a strong positive correlation: \textit{$m^\star$ increases with $k$ at a steady compression ratio $\rho$ around $2$}.
The larger Qwen2.5-VL-72B model consistently demonstrates a superior compression ratio compared to both the 7B model and GPT-4.1-mini. The approximately \textbf{linear relationship suggests a predictable trade-off} between visual budget and text compression capacity.

We next compare wall-clock generation time (Table~\ref{tab:latency}).
For GPT-4.1-mini the vision processing adds a modest $<1.5$s overhead,\footnote{\textbf{Not} exact on model due to API overhead. See Apendix~\ref{sec:Images}.}
whereas for Qwen2.5-VL the shorter sequence length outweighs that cost,
yielding 25–45\% faster inference. Across accuracy and latency, converting long textual contexts into
images yields substantial token savings without
sacrificing RULER retrieval performance, demonstrating a simple yet effective
way to reduce inference cost on large-context tasks.

\begin{table*}[h!]
  \small
  \centering
  \begin{subtable}[t]{0.58\textwidth}
    \centering
    \vspace{0pt} 
    \setlength{\tabcolsep}{4pt}
    \begin{tabular}{lcccccc}
      \toprule
      \multirow{2}{*}{\textbf{Model}} &
      \multicolumn{3}{c}{\textbf{Text–Image (hybrid)}} &
      \multicolumn{3}{c}{\textbf{Text-only}}\\
      \cmidrule(lr){2-4}\cmidrule(lr){5-7}
      & Image size & $k$ & Acc.\ (\%) & $m^\star(k)$ & Acc.\ (\%) & $k/m^\star(k) \!\downarrow$ \\
      \midrule
      \textsc{gpt}
      & $600{\times}800$  & 783  & 99 & 1,272.4 & 100 & 0.61 \\
      & $600{\times}1000$ & 998  & 97 & 1,752.5 & 100 & 0.57 \\
      & $750{\times}1000$ & 1,258& 98 & 2,352.2 & 100 & 0.53 \\[2pt]
      \textsc{qwen}
      & $600{\times}800$  & 635  & 98 & 1,289.6 & 100 & 0.49 \\
      & $600{\times}1000$ & 782  & 99 & 1,769.7 & 100 & 0.44 \\
      & $750{\times}1000$ & 998  & 98 & 2,369.4 & 100 & 0.42 \\
      \bottomrule
    \end{tabular}
    \caption{\textbf{RULER accuracy and token statistics.} 
             $k$ is the visual token count, $m^\star(k)$ the maximum \textit{text-token tolerance}, and $k/m^\star(k)$ the relative token footprint.
             Text-image input reduces the decoder tokens by 38–58\%.
             }
    \label{tab:nihs_accuracy}
  \end{subtable}%
  \hfill
  \begin{subtable}[t]{0.38\textwidth}
    \centering
    \vspace{0pt} 
    \begin{tabular}{lccc}
      \toprule
      \textbf{Model} & $k$ & Time$_{\text{img}}$ (s)$\downarrow$ & Time$_{\text{text}}$ (s)$\downarrow$ \\
      \midrule
      \textsc{GPT}$^*$
      & 783  & 1.29 & 0.60 \\
      & 998  & 1.75 & 0.61 \\
      & 1,258& 1.95 & 0.58 \\[2pt]
      \textsc{qwen}
      & 635  & 2.81 & 3.61 \\
      & 782  & 3.04 & 4.16 \\
      & 998  & 3.35 & 5.09 \\
      \bottomrule
    \end{tabular}
    \caption{\textbf{End-to-end latency.}
             Hybrid inputs add modest overhead for GPT but
             \emph{reduce} total time for Qwen thanks to smaller decoder
             contexts.
             $^*$measured from API responses; see Appendix~\ref{sec:Images}. 
             }
    \label{tab:latency}
  \end{subtable}
  \caption{RULER S-NIAH long-context retrieval accuracy, text-as-image compression statistics, and model latency. 
  \textsc{gpt} refers to GPT-4.1-mini and \textsc{Qwen} denotes Qwen2.5-VL-72B-Instruct.
  }
\end{table*}

\begin{table*}[h]
\centering
\small
\setlength{\tabcolsep}{5.5pt}
\begin{tabular}{llrrrrrr}
\toprule
\textbf{Model} & \textbf{Method} &
\textbf{Remaining} $k$\,$\!\downarrow$ &  \textbf{BERTScore} &\textbf{ROUGE-L} & \textbf{ROUGE-1} & \textbf{ROUGE-2} & \textbf{ROUGE-L\textsubscript{sum}} \\
\midrule
\multirow{4}{*}{\textsc{GPT}}
& Baseline (text-only)   & $m{=}693$ &  \textbf{86.25} &\textbf{16.26} & \textbf{23.78} & \textbf{8.60} & \textbf{18.91}\\
& Text-as-image (ours)      & 225 $(\!{-}67\%)$ &  85.33 &15.31 & 21.98 & 7.40 & 17.75\\
& Select-Context    & 295 $(\!{-}57\%)$ &  85.01 &12.79 & 18.82 & 5.30 & 15.71\\
& LLMLingua-2       & 265 $(\!{-}62\%)$ &  85.25 &13.75 & 20.42 & 7.00 & 16.60\\[2pt]
\multirow{4}{*}{\textsc{Qwen}}
& Baseline (text-only)   & $m{=}726$ &  \textbf{86.37} &\textbf{17.70} & \textbf{25.18} & \textbf{9.47} & \textbf{20.77}\\
& Text-as-image (ours)     & 279 $(\!{-}62\%)$ &  85.64 &15.53 & 23.28 & 7.54 & 19.16\\
& Select-Context    & 181 $(\!{-}75\%)$ &  84.60 &13.40 & 20.36 & 5.13 & 17.63\\
& LLMLingua-2       & 258 $(\!{-}64\%)$ &  85.46 &15.32 & 22.95 & 7.09 & 18.94\\
\bottomrule
\end{tabular}
\caption{CNN/DailyMail document summarization with token compression
baselines. \textsc{gpt} refers to GPT-4.1-mini and \textsc{Qwen} denotes Qwen2.5-VL-72B-Instruct.
$m$ is the uncompressed text length, $k$ the remaining
decoder tokens after compression.  Percentage shows token reduction
relative to $m$.  At \emph{similar} compression rates, rendering the
document as an image beats both token-level baselines on every metric.}
\label{tab:cnn_baselines}
\end{table*}

\subsection{Document–Level Summarization}
\label{sec:summarisation}

While \textsc{RULER} stresses a model’s ability to \emph{retrieve} a
single item from a long context, it is \emph{not} expressly designed
as a compression benchmark—every token in the haystack is, by
construction, irrelevant to the answer.  To gauge how well
text–as-image compression fares on a genuine \emph{compression} task, we
turn to CNN/DailyMail long–document summarization, where \emph{all}
input sentences may contribute to the final summary.

\paragraph{Setup.}
We compare our approach against two widely used token–pruning
techniques that operate purely in the text modality: (1) \textbf{Select–Context}~\citep{li-etal-2023-compressing}: keeps tokens
          whose self-information (estimated by a small LLM) exceeds a
          learned threshold.
(2) \textbf{LLMLingua-2}~\citep{pan-etal-2024-llmlingua}: trains a
          Transformer encoder to predict, token by token, whether to
          retain or discard.

We compute the average input length in CNN/DailyMail and use it to set the image resolution in our text-to-image pipeline, yielding visual token inputs at roughly half the original context length—the optimal ratio identified in previous section. This image size is applied consistently across the task. For fair comparison, baselines are configured with the same decoder token compression ratio. Summary quality is evaluated with \textsc{ROUGE} \citep{lin-2004-rouge} and \textsc{BERTScore} \citep{zhang2020bertscoreevaluatingtextgeneration}.

\paragraph{Results and Discussion.}
The evidence in Table~\ref{tab:cnn_baselines} shows that, even though
text–image compression was \emph{not} tailored for summarization, it
yields stronger summaries than two specialized pruning methods while
retaining only $\sim$$40\%$ of the original tokens.  We stress that
the aim of this paper is to \emph{characterize} the compression
capacity of visual inputs, not to introduce yet another SOTA
compression model.  Nevertheless, the unexpectedly strong results
suggest that our simple rendering trick is a competitive—and
orthogonal—alternative to learned token-selection approaches.  Future
work could further combine text token pruning \emph{before} rendering,
stacking the benefits of both paradigms.

\section{Conclusion}
\label{sec:conclusion}

Our primary goal is to answer the question:
\emph{``How many tokens can be saved by replacing text with an
image without harming downstream performance?''} 
By converting long
contexts into visual form, we achieved nearly two-fold reductions in
decoder token count while preserving 
task accuracy on both retrieval (\textsc{RULER S-NIAH}) and generation
(CNN/DailyMail summarization) benchmarks.  The approach is model- and
task-agnostic, requires no parameter updates, and can even lower
latency on large decoders. Our findings suggest several directions for future work: (i) applying token-level pruning before visual rendering to further compound compression gains; and (ii) expanding the approach to other domains (e.g. math) where most prompt tokens are critical and thus difficult to prune at the token level. We hope this study will spark broader exploration of modality shifting as a complementary axis for scaling the usability and efficiency of Large language models.

\section*{Limitations}
Despite showing promising token savings on short to medium context scenarios, our work has not yet fully evaluated the impact of text-as-image prompting on \textit{extremely large} contexts that span tens of thousands of tokens or more. Real-world applications such as document analysis or in-depth conversational histories may require specialized approaches (e.g., chunking, retrieval) to ensure reliable performance at these larger scales. Furthermore, our experiments focus on a limited number of benchmarks, leaving open questions about performance on other domains (e.g., medical, legal) and tasks (e.g., coding, translation).

\section*{Acknowledgment}

We thank the Google Gemma Academic
Program for their partial support of Jiawei Zhou
and for providing computational resources.

\bibliography{custom}

\clearpage

\appendix
\section*{Appendix}



\section{Results Across Context Lengths and Image Sizes on \textsc{Ruler}}
\label{sec:Images}

A central challenge is how to balance visual and textual tokens under a fixed sequence length. In many patch-based vision encoders used by VLMs such as Qwen, higher image resolution produces more patches, and thus more visual tokens, which offsets efficiency benefits of text-as-image processing for token reduction. Conversely, lowering resolution reduces token usage, but risks discarding semantically important visual details. This tradeoff becomes especially critical in long-context settings such as retrieval-style reasoning or the \textsc{Ruler} benchmark, where even small shifts in token allocation can lead to large changes in accuracy.

Our experiments demonstrate that model accuracy is highly sensitive to the effective resolution of image inputs, which directly controls the number of visual tokens generated. We observe a consistent trend across benchmarks: \textit{when visual tokens account for roughly one half of the total context window, performance remains nearly indistinguishable from that achieved with the original, uncompressed textual context.} This $\tfrac{1}{2}$ allocation emerges as a near-optimal operating point, striking a balance between preserving visual fidelity and maintaining sufficient capacity for textual information. The finding aligns with broader evidence of redundancy in long-context models, where moderate compression often preserves accuracy despite substantial reductions in token count.

\begin{figure*}[t]
    \centering
    \includegraphics[width=\textwidth]{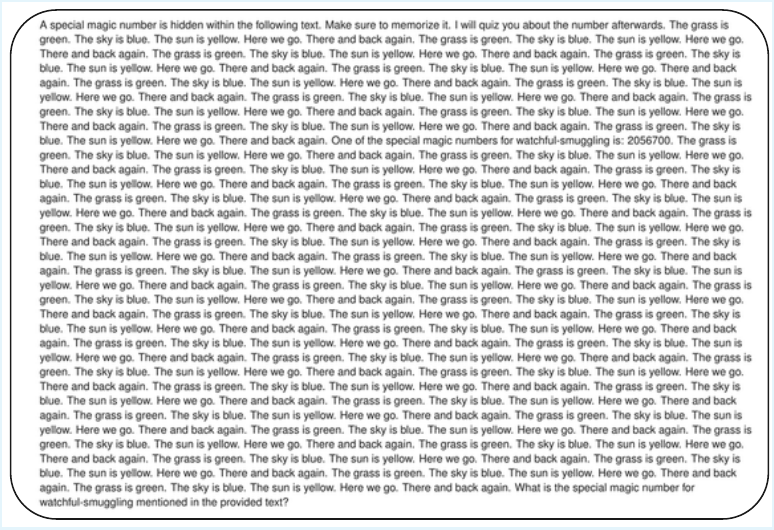}
    \caption{Rendered image input for the \textsc{Ruler} task at context length 1500 ($600\times800$ image resolution). Here there is almost \textbf{no} accuracy degradation. This example illustrates how textual sequences are converted into rasterized images for multimodal processing.}
    \label{fig:1500}
\end{figure*}

\begin{figure*}[t]
    \centering
    \adjustbox{max width=\textwidth, max height=0.95\textheight}{
        \includegraphics{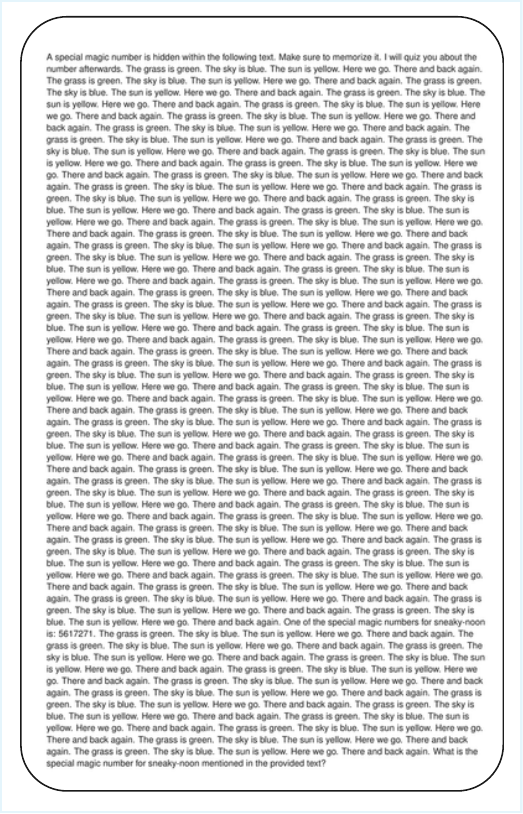}
    }
    \caption{Rendered image input at context length 2000 ($600\times1000$ image resolution). Here there is almost \textbf{no} accuracy degradation. The figure demonstrates scaling of resolution while preserving readability and model performance.}
    \label{fig:2000p}
\end{figure*}

\begin{figure*}[t]
    \centering
    \includegraphics[width=\textwidth]{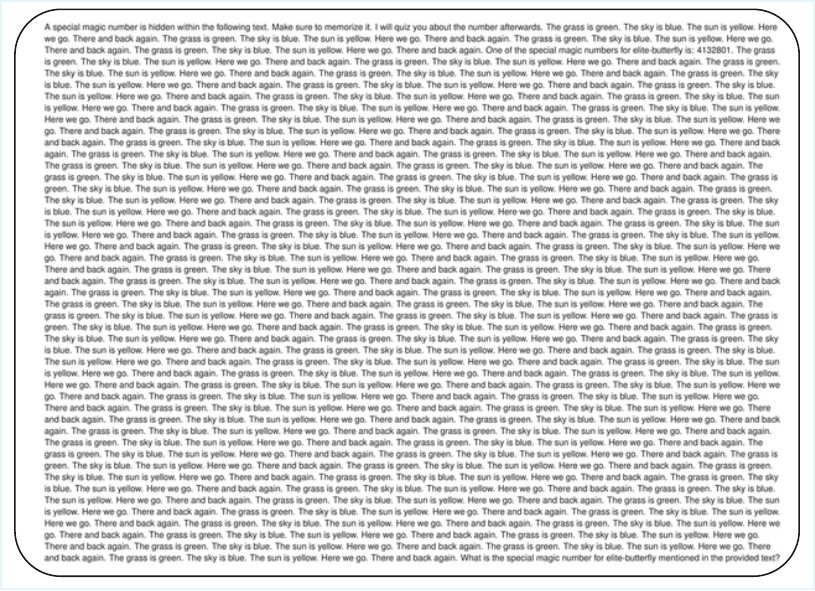}
    \caption{Rendered image input at context length 2500 ($750\times1000$ image resolution). Here there is almost \textbf{no} accuracy degradation. Increased resolution produces more visual tokens while maintaining comparable visual fidelity and model performance.}
    \label{fig:2500p}
\end{figure*}

To illustrate, Figure~\ref{fig:1500}–\ref{fig:2500p} present representative cases from the \textsc{Ruler} needle-in-a-haystack task: at a context length of 1500 with an image resolution of $600\times800$ pixels (Figure~\ref{fig:1500}), at 2000 with $600\times1000$ pixels (Figure~\ref{fig:2000p}), and at 2500 with $750\times1000$ pixels (Figure~\ref{fig:2500p}). In all three cases, the model successfully recovers nearly \emph{all} embedded image information, achieving accuracy close to 100\%.

\begin{figure*}[t]
    \centering
    \adjustbox{max width=\textwidth, max height=0.95\textheight}{
        \includegraphics{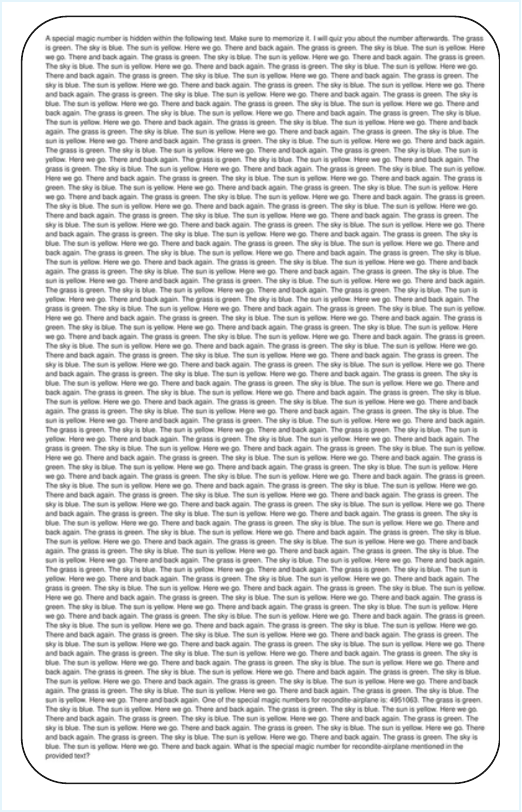}
    }
    \caption{Rendered image input at context length 3000 ($600\times1000$ image resolution). Here accuracy \textbf{degrades} substantially. This setting illustrates the tradeoff between tolerable text token budget and image resolution to maintain model performance.}
    \label{fig:3000p}
\end{figure*}

In contrast, when the context length is extended to 3000 tokens with a $600\times1000$ image (Figure~\ref{fig:3000p}), accuracy degrades substantially despite the increase in available tokens. This illustrates that recognition accuracy depends not only on the absolute context length but also on preserving a balanced allocation between textual and visual tokens. Simply enlarging the context window is therefore insufficient to ensure stable performance.

Collectively, these results indicate that tuning image resolution relative to the available context budget is critical for multimodal reasoning. The \textbf{observed $\tfrac{1}{2}$ allocation} (or compression ratio $\rho$ of 2) emerges as a practical heuristic for balancing efficiency and fidelity, though further validation is required to assess its robustness across tasks, datasets, and model architectures.

\paragraph{Inference Latency Analysis}


To measure the inference latency of text-only input to the LLM decoder and text-image hybrid input to the full multimodal model (shown in Table~\ref{tab:latency}), we run the full RULER S-NIAH test set and report the average wall-clock time per sample. No batching is used. For the proprietary GPT-4.1-mini, latency is measured from API response times. For Qwen2.5-VL-72B-Instruct which is open sourced, inference is performed on 8 Nvidia RTX A6000 GPUs using the standard Hugging Face implementation without batching, vLLM, or other speedups.

Note that API response times for GPT-4.1-mini may \textit{not reflect pure model latency}, as they include system overhead such as request queuing, routing, network latency, and data transmission. Since image payloads are larger than text, this overhead is likely greater for text-as-image inputs, partially explaining the higher latency reported in Table~\ref{tab:latency}.

\section{Results on Qwen2.5-VL-7B-Instruct}
\label{appendix:qwen-7b}

\begin{figure}[t]
    \centering
    \includegraphics[width=\linewidth]{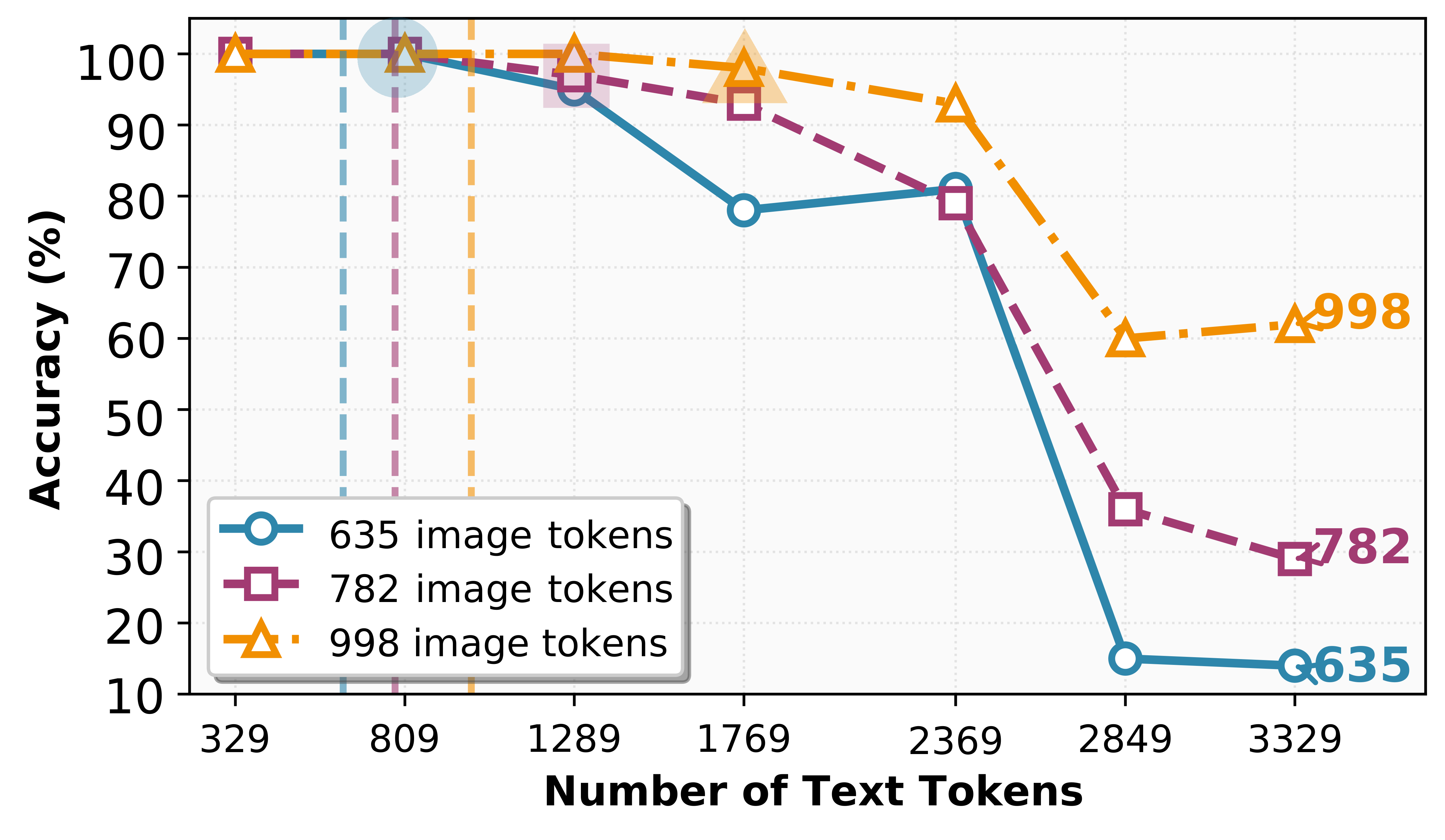}
    \caption{\textbf{Qwen2.5-VL-7B-Instruct: accuracy vs.\ text-token size ($m$) with fixed visual tokens ($k$)}. Each curve varies the amount of rendered text for a fixed visual token size. Compared to the 72B version (Figure~\ref{fig:qwen_curve_sub}), the smaller 7B model exhibits a much steeper performance degradation as text density increases, indicating that model scale is a critical factor for effective visual text processing.}
    \label{fig:qwen_7b_curve}
\end{figure}

To understand the role of model scale in visual text compression, we replicate the long-context retrieval experiment from Section~\ref{subsec:long-context-retrieval} on the much smaller \textbf{Qwen2.5-VL-7B-Instruct} model.\footnote{\url{https://huggingface.co/Qwen/Qwen2.5-VL-7B-Instruct}.} The results, plotted in Figure~\ref{fig:qwen_7b_curve}, show that while the general trend of performance degradation holds, the 7B model is significantly more sensitive to text density than its 72B counterpart.

The text-token tolerance of the 7B model is substantially lower across all corresponding visual token budgets ($k$). For instance, with $k=998$ visual tokens, the 72B model maintains over 97\% accuracy up to nearly 2,400 text tokens, whereas the 7B model's accuracy drops below 95\% after only 2,000 text tokens. This highlights that larger model scale provides greater capacity to robustly process and reason over densely packed textual information presented visually, making it a critical factor for achieving high compression ratios without performance loss.

\section{BABILong 1k Benchmark}
\label{app:babilong}

To further validate our findings, we evaluate on the \textsc{BABILong} benchmark~\citep{kuratov2024babilongtestinglimitsllms}, which is specifically designed to test the limits of long-context reasoning in LLMs. The benchmark extends the classical bAbI tasks to much longer contexts, where the relevant information must be retrieved from sequences containing up to 1k distractor tokens. This makes it well suited for assessing whether multimodal token allocation strategies remain effective in long-context scenarios.  

We follow the standard setup of the \textsc{BABILong} 1k variant, which consists of ten subsets (QA1–QA10) covering a range of reasoning tasks such as single- and multi-supporting fact retrieval, coreference, and induction. For this experiment, we use \texttt{gemini-2.5-flash-preview-04-17}~\citep{comanici2025gemini25pushingfrontier}.  

Since Gemini is closed-source, we estimate visual token usage from the input/output token statistics returned by the API. Specifically, we record both text and image token counts for each query, enabling us to analyze how token allocation is distributed between modalities.  

Table~\ref{tab:gemini2p5_babilong_1k} shows that across QA1–QA10, performance is stable when image tokens constitute \textbf{roughly half} the count of the original total context text tokens. In this configuration, the model achieves strong and well-aligned accuracies for both text (0.91) and image (0.83), confirming that the \textbf{$\tfrac{1}{2}$ allocation rule} observed in the \textsc{Ruler} experiments generalizes to BABILong as well. This provides additional evidence that balancing image and text tokens at a near-equal ratio offers a practical operating point for multimodal long-context reasoning.

\begin{table*}[h]
\centering
\small
\begin{tabular}{lcccccc}
\toprule
\textbf{Task} & \textbf{TxtAcc} & \textbf{ImgAcc} & \textbf{TxtIn} & \textbf{TxtOut} & \textbf{ImgIn} & \textbf{ImgOut} \\
\midrule
QA1   & 1.00 & 0.96 & 846.9 &  9.0 & 440.0 &  9.1 \\
QA2   & 0.85 & 0.65 & 877.0 &  7.0 & 466.0 &  7.0 \\
QA3   & 0.87 & 0.59 & 946.2 & 10.0 & 531.0 & 10.0 \\
QA4   & 1.00 & 0.98 & 810.5 &  1.1 & 399.0 &  1.1 \\
QA5   & 0.92 & 0.89 & 881.0 &  1.0 & 465.4 &  1.0 \\
QA6   & 0.99 & 0.96 & 829.0 &  1.0 & 430.0 &  1.0 \\
QA7   & 0.54 & 0.48 & 863.1 &  1.0 & 451.0 &  1.0 \\
QA8   & 0.99 & 0.95 & 862.7 &  1.1 & 468.0 &  1.1 \\
QA9   & 1.00 & 0.96 & 827.7 &  1.0 & 425.0 &  1.0 \\
QA10  & 0.98 & 0.92 & 875.1 &  1.0 & 466.0 &  1.0 \\
\midrule
\textbf{Avg} & 0.91 & 0.83 & 861.9 &  3.3 & 454.1 &  3.3 \\
\bottomrule
\end{tabular}
\caption{\textbf{Gemini-2.5-flash-preview-04-17 on \textsc{BABILong} 1k.} ``TxtAcc'' is accuracy for text input. ``ImgAcc'' is accuracy for image input. ``TxtIn'' and ``TxtOut'' are the average text input/output tokens. ``ImgIn'' and ``ImgOut'' are the average image input/output tokens.}
\label{tab:gemini2p5_babilong_1k}
\end{table*}

\section{ConTexImage: A Text-to-Image Conversion Pipeline}
\label{sec:ConTexImage}

To standardize our experiments, we require a consistent way to convert textual sequences into image inputs of controlled resolution. 
We therefore design a lightweight text-to-image pipeline, which we term \textbf{ConTexImage} (Algorithm~\ref{alg:conteximage}). 
ConTexImage renders arbitrary text into rasterized images while automatically adjusting font size to achieve a target content density. 
This ensures that the generated images preserve readability, maintain consistent tokenization patterns, and remain comparable across different resolutions.

The pipeline consists of three main stages:
\begin{enumerate}
    \item \textbf{Preprocessing:} Input text is normalized by replacing typographic symbols (e.g., curly quotes, dashes, ellipses) and escaping LaTeX special characters. This step guarantees compatibility with the rendering backend.
    \item \textbf{LaTeX Rendering:} The cleaned text is embedded into a minimal \LaTeX{} document and compiled using \texttt{tectonic}\footnote{\url{https://github.com/tectonic-typesetting/tectonic}.}. The output PDF page is rasterized into an image at a specified DPI and then resized to the target resolution (e.g., $600\times800$, $600\times1000$).
    \item \textbf{Adaptive Font Optimization:} To maximize visual fidelity, the algorithm searches over candidate font sizes and evaluates the proportion of the image occupied by text (\emph{fill ratio}). The largest font size that meets a pre-defined target fill ratio (default 0.8) is selected, ensuring both legibility and balanced whitespace.
\end{enumerate}

The resulting images are consistent across different contexts and allow us to precisely control the number of image tokens relative to text tokens. More details are illustrated in Algorithm~\ref{alg:conteximage}.


\renewcommand{\lstlistingname}{Algorithm}

\begin{figure*}[t]
\centering
\begin{lstlisting}[language=Python, 
                   caption={\textbf{ConTexImage}: A Context-Aware Text-to-Image Pipeline}, 
                   label={alg:conteximage},
                   frame=lines]
def escape_latex_special_chars(text):
    """Replace LaTeX special characters with safe tokens."""
    escape_map = {"\\": r"\textbackslash{}", "&": r"\&", "%": r"\%",
                  "$": r"\$", "#": r"\#", "_": r"\_", "{": r"\{", "}": r"\}",
                  "~": r"\textasciitilde{}", "^": r"\textasciicircum{}"}
    for k, v in escape_map.items():
        text = text.replace(k, v)
    return text

def text_to_image(text, width, height, dpi=300, target_fill_ratio=0.7):
    # Step 1: Preprocessing
    text = normalize_typography(text)        # replace curly quotes, dashes, ellipses
    text = escape_latex_special_chars(text)  # escape special symbols for LaTeX

    # Step 2: Font size search
    best_image, best_ratio = None, 0
    for font_size in candidate_font_sizes(descending=True):
        tex_doc = build_latex_template(text, font_size, width, height, margin=10)
        pdf = compile_with_tectonic(tex_doc)                  # LaTeX -> PDF
        image = convert_pdf_to_image(pdf, dpi=dpi,
                                     resize=(width, height))  # PDF -> raster
        ratio = calculate_fill_ratio(image)                   # bounding box occupancy

        if ratio > best_ratio:     # update best so far
            best_ratio, best_image = ratio, image
        if ratio >= target_fill_ratio:
            break

    return best_image

def generate_images(dataset, output_dir, width, height, dpi=300):
    """Batch conversion for all documents in dataset."""
    for doc in dataset:
        text = doc["input"]
        img = text_to_image(text, width, height, dpi=dpi)
        save_image(img, path=output_dir + f"/{doc['doc_id']}.png")
\end{lstlisting}
\end{figure*}

\renewcommand{\lstlistingname}{Listing} 

\end{document}